# Text Classification Based on Knowledge Graphs and Improved Attention Mechanism

Siyu Li, Lu Chen, Chenwei Song, Xinyi Liu
School of Information Science and Engineering, Chongqing Jiaotong University
{lsy20, chenlu, chenweisong, xyliu}@mails.cqjtu.edu.cn

**Abstract:** To resolve the semantic ambiguity in texts, we propose a model, which innovatively combines a knowledge graph with an improved attention mechanism. An existing knowledge base is utilized to enrich the text with relevant contextual concepts. The model operates at both character and word levels to deepen its understanding by integrating the concepts. We first adopt information gain to select import words. Then an encoder-decoder framework is used to encode the text along with the related concepts. The local attention mechanism adjusts the weight of each concept, reducing the influence of irrelevant or noisy concepts during classification. We improve the calculation formula for attention scores in the local self-attention mechanism, ensuring that words with different frequencies of occurrence in the text receive higher attention scores. Finally, the model employs a Bi-directional Gated Recurrent Unit (Bi-GRU), which is effective in feature extraction from texts for improved classification accuracy. Its performance is demonstrated on datasets such as AGNews, Ohsumed, and TagMyNews, achieving accuracy of 75.1%, 58.7%, and 68.5% respectively, showing its effectiveness in classifying tasks.

**Keywords:** text classification; knowledge graph; Natural Language Processing; improved attention mechanism

## 0 Introduction

Social networks have surpassed traditional media to become new information gathering places, and have influenced the information dissemination pattern of society at a very fast speed [1]. How to classify these short texts accurately is a key technology in the field of Natural Language Processing, NLP). Because these short texts are short in length, lack of contextual information, colloquial in content, with many characteristic attributes and large noise, it is an urgent problem to accurately extract text features and classify short texts with appropriate classification models.

In the field of text classification, general text representation methods are divided into explicit representation and implicit representation. For explicit representation, people generally create effective features from knowledge base, part-of-speech tagging, syntactic analysis [2] and other aspects, and express short texts as sparse vectors, and each dimension is an explicit feature. Although the explicit representation of the text is easy to be understood, it often ignores the context of the short text and cannot capture the deep semantic information. In addition, there are problems of sparse data. For example, when the entity features don't exist in the knowledge base, you can't get any of its features, and the explicit representation won't work. At present, the implicit representation method of implicit text is more common in deep learning. By training word vectors, each word is mapped into a dense vector [3], and the short text is represented by a word vector matrix. Because the word vectors contain semantic information, the neural network model can obtain richer semantic information from the context and promote the understanding of the short text by the neural network model. However, there are still some shortcomings in the implicit representation method, such as the short text {The Bulls Won the NBA Championship}, in which Bulls is the name of a basketball team. However, the model input by word vector may not capture this information and regard it as an animal or a new word, resulting in unsatisfactory classification effect. There are some problems in simply using explicit or implicit text representation, so combining the two methods, using a rich knowledge base to enrich the prior knowledge of short texts, obtaining the concept set of short texts, and then mapping the short texts and concept sets into a word vector matrix, so that the model can learn more comprehensive and deeper semantics and improve the classification ability.

In this paper, a neural network model integrating knowledge graph and attention mechanism is constructed. The knowledge graph is integrated with the short text classification model, and the concept set of short text is obtained from the existing knowledge base as input, so as to obtain the prior knowledge in the text. Based on this, the attention mechanism is introduced to calculate the relevance of each concept to the concept set and the short text, and the attention weights of the two are weighted and fused to obtain the final weight of each concept, so as to increase the weight of the related concepts, and to make the model more discriminative in its classification effect.

## 1 Related Work

With the development of deep learning, more and more scholars use deep learning method to study text classification, and its performance is better than traditional methods in most tasks, which greatly promotes the development of text classification [4-10]. In 2016, Microsoft Research released Concept Mapping. Concept Mapping is a large-scale knowledge graph system that captures a large amount of commonsense knowledge by learning from hundreds of millions of web pages and years of accumulated search diary data. A concept map is represented as a triad of instances, concepts, and relationships. The relationship between instances and concepts is IsA, e.g., the triad (apple, fruit, IsA) indicates that apple is a fruit. Literature [11] proposes six methods for instance conceptualization based on concept mapping, and provides the corresponding API function calls on the official website of Microsoft Knowledge Graph.

In 2017, WANG et al. [12] proposed a CNN short text categorization model integrating concept mapping, by combining the a priori knowledge of short text obtained in advance from the knowledge base with the text features extracted by CNN, the shortcomings of short text categorization that lacks contextual information can be solved to a certain extent, and excellent results have been obtained in five public datasets. It can be seen that the combination of knowledge graph and deep learning can solve the problem of the lack of context in short text, and the model incorporating knowledge graph can obtain additional information beyond word vectors, and has better performance in text categorization tasks. In addition, many methods [13-16] are proposed for knowledge graph reasoning which can further help build the connection between concepts. Although the neural network model incorporating the knowledge graph has better performance, there are still some problems. For example, in the above article, "The Bulls won the NBA championship", although the concepts of team and animal can be obtained from the knowledge base, it is obvious that animal is an inappropriate concept in the text, and the noise will affect the results of text categorization. In addition, for example, if we input the text {Steve Jobs is one of the cofounders of Apple,} , we can retrieve the concepts of Steve Jobs as an entrepreneur and as a person from the Knowledge Base, but although both concepts are correct, it is obvious that the concept of entrepreneur should be given more weight in the short text.

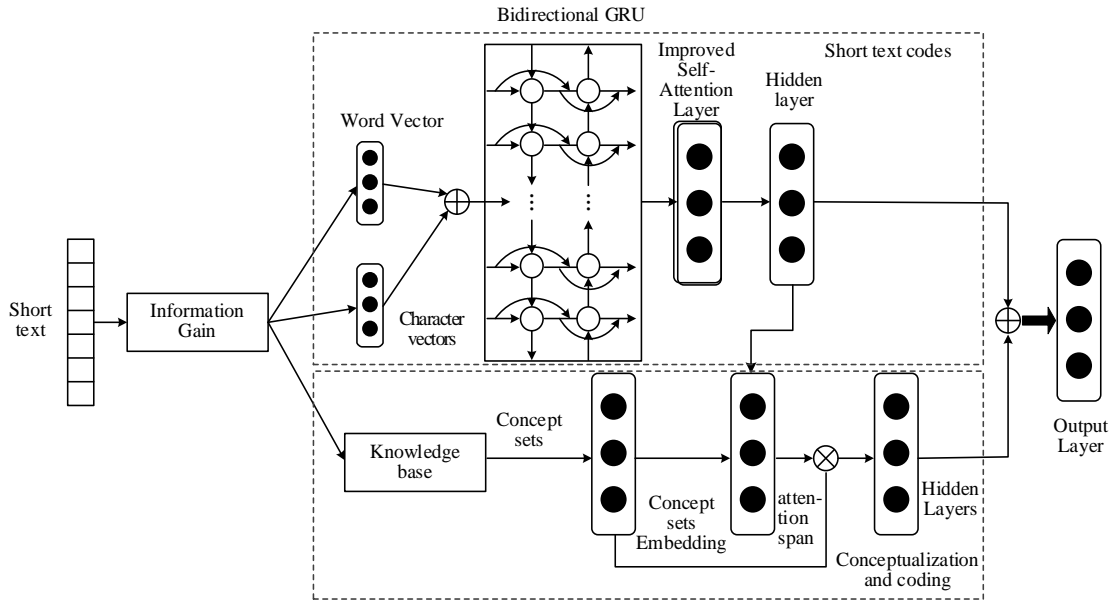

Fig.1 Structure of text classification model.

In order to solve the above problems, this paper introduces the attention mechanism, and at the same time, draws lessons from the Transformer model, and puts forward a kind of attention-gated circulating unit network with knowledge map. By calculating the attention weight of the short text and the concepts in its concept set, the concepts closely related to the short text are given higher weight. In the above example, {Steve Jobs is one of the co-founders of Apple}, Steve Jobs' concepts of entrepreneur and individual increase the weight of the concept of entrepreneur and decrease the weight of the concept of individual, which makes the text classification model more discriminating.

**2 Knowledge Enhanced Text Classification Model**

In this paper, a knowledge enhanced attention bi-GRU (Keat-GRU) is proposed, which integrates knowledge map, attention mechanism and two-way GRU.

1) Short text coding: using character vectors and word vectors as input, extracting short text features through Bi-GRU, and weighting important text information by multi-head self-attention layer to obtain short text features.

2) Conceptual coding: by calling the API of Microsoft Concept Map, the concept set of short text is obtained and vectorized, and the weight of the concepts closely related to the short text in the concept set is improved by calculating the Attention with the feature vector of the short text, and finally the characteristics of the concept set are obtained

2.1 Information Gain

Information gain in probability refers to the degree of reduction of information complexity, that is, the uncertainty of information, under a condition, which is the difference between the information extract and the condition. Entropy is a measure of uncertainty or random variable, suppose a random variable

$$X = \{x_1, x_2, \cdots, x_n\} \quad (1)$$

The probability distribution is $p(x)$, then the entropy of the random variable is

$$H(x) = -\sum_{x \subset X} p(x) \log p(x) \quad (2)$$

In text categorization, information gain evaluates the importance of a feature word by the amount of information it can provide to the whole classification, and is the difference between the extract of the text without features and the extract of the text after adopting the features, and the formula for IG is

$$\begin{aligned} \text{IG}(x) = &H(C) - H(C \mid x) = \\ &-\sum_{i=1}^{m} P(C_i) \log P(C_i) + \\ &P(x) \sum_{i=1}^{m} P(C_i \mid x) \log P(C_i \mid x) + \\ &P(\bar{x}) \sum_{i=1}^{m} P(C_i \mid \bar{x}) \log P(C_i \mid \bar{x}) \end{aligned} \quad (3)$$

Where: $P(C_i) - C_i$ the frequency of occurrence of the class document in the corpus; $P(x)$ —Frequency of documents with feature $x$; $P(C_i \mid x)$ —The probability that a document with feature $x$ belongs to class $C_i$; $P(\bar{x})$ —Probability of documents without feature $x$; $P(C_i \mid \bar{x})$ — Documents without feature $x$ belong to $C_i$. $M$ —Number of categories.

2.2 Short Text Encoding

Given the input short text word sequence $\{x_1, x_2, \cdots, x_n\}$, the character-level vector $e_i^{W_c}$ of the word is trained, which is calculated as shown in Eq. (1):

$$e_i^{W_c} = \max_{1 \leqslant j \leqslant L} \left( W_{\text{CNN}}^T \begin{bmatrix} e^c\left(c_{j-\frac{kc-1}{2}}^2\right) \\ \vdots \\ e^c\left(c_{j-\frac{kc-n}{2}}^2\right) \end{bmatrix} + b_{\text{CNN}} \right) \quad (4)$$

Among them, $W_{\text{CNN}}$ and $b_{\text{CNN}}$ are training parameters, $ke$ represents the convolution kernel size, and $max$ represents the maximum pooling operation.

The model then maps the word $x_i$ to the word vector $e^w$:

$$e_i^w = E(x_i) \quad (5)$$

The word vector is spliced with the character vector:

$$E_i = \text{Concat}\left(e_i^w; e_i^{W_c}\right) \quad (6)$$

Finally, the word vector matrix $E = [E_1, E_2, \cdots, E_n]$ is obtained as the input of Bi-GRU. The forward GRU reads the input sequence $(E_1 \sim E_n)$ in the normal order, while the reverse GRU reads the input sequence $(E_n \sim E_1)$ in the reverse order. The input vector E_i at each time t is calculated by the gated loop unit, and the forward hidden state $(\vec{h}_1, \vec{h}_2, \cdots, \vec{h}_t)$ at each time is obtained.

$(\overleftarrow{h}_1, \overleftarrow{h}_2, \cdots, \overleftarrow{h}_t)$, connecting the forward hidden state $\vec{h}_j$ at each moment with the reverse hidden state $\overleftarrow{h}_j$ at the corresponding moment to obtain the hidden state at that moment:

$$h_j = [\vec{h}_j \parallel \overleftarrow{h}_j]^T \quad (7)$$

Then, the hidden state $h_i$ at each moment is input into the self-Attention layer, and the input words at each time step are weighted according to the attention calculation, so that the important words get higher weight. The attention calculation is defined as:

$$\text{Attention}(Q, K, V) = \text{softmax}\left(\frac{QK^T}{\sqrt{d_k}}\right)V \quad (8)$$

Among them, $Q$ represents the query when $Attention$ is executed once, $K$ represents the similarity corresponding to the value and used to calculate the similarity with the query as the basis for $Attention$ selection, and $V$ represents the data that is noticed and selected. $Q \in \mathbb{R}^{n \times d_k}, K \in \mathbb{R}^{m \times d_k}, V \in \mathbb{R}^{m \times d_v}$, and the input includes query and key in $d_k$ dimension and value in $d_v$ dimension.

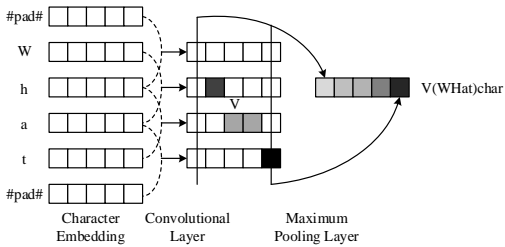

Fig.2 Structure of character level convolution neural network

The Multi-head Attention operation structure is shown in Fig. 3. The computation is defined as follows:

$$head_i = \text{Attention}(QW_i^Q, KW_i^K, VW_i^V) \quad (9)$$

$$\text{MultiHead}(Q, K, V) =$$

$$\text{Concat}(head_1, head_2, \cdots, head_h) \quad (10)$$

$$a_i = \text{MultiHead}(h_i, h_i, h_i) \quad (11)$$

$h$ different linear transformations are used to map the key, value and query of $d_{\text{model}}$ dimension into $d_k$ dimension, $d_k$ dimension and $d_v$ dimension respectively, and then the output of $h \times d_v$ dimension is calculated, spliced, and finally the final output is obtained by a linear transformation. $h_i$ is the input sequence, that is, the hidden state of Bi-GRU layer output, and its purpose is to calculate the attention within the input sequence and find the connection within the sequence.

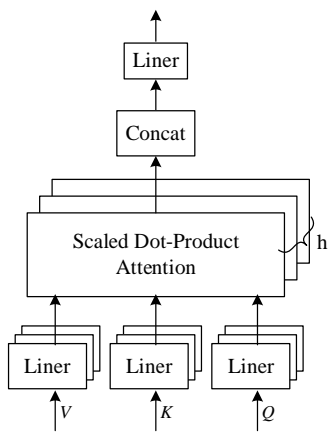

Fig.3 Operation structure of Multihead Attention.

By calculating the attention weight $a_t$ from the attention layer, the hidden state $h_t$ at time $t$ output by Bi-GRU is weighted and averaged:

$$h'_t = \sum_{t=1}^{T} a_{t'} h_t \quad (12)$$

Finally, the characteristic matrix $h' \in \mathbb{R}^{n \times 2u}$ is output.

2.3 Short Text Conceptualization Encoding

Text conceptualization needs to go through existing knowledge bases such as Yago [17] and Microsoft Concept Graph. In this paper, the Concept Graph knowledge map released by Microsoft is used to conceptualize the short text and obtain the concept set related to the text. Get the concept set $C = (c_1, c_2, \cdots, c_m)$ of each short text from the knowledge base, and $c_i$ represents the concept vector in the $i$th concept set. In order to increase the weight of key concept vectors and reduce the influence of concept vectors irrelevant to the short text on the results, firstly, the feature matrix $h' \in \mathbb{R}^{n \times 2u}$ of the short text is transformed into the feature vector $q \in \mathbb{R}^{2u}$ through the maximum pooling layer, and then the relationship weight between the $i$-th vector in the concept set and its feature vector $q$ is calculated by the attention-grabbing mechanism:

$$\alpha_i = \text{softmax}\left(v_1^T \tanh(W_1 \times \text{concat}[c_i; q] + b_1)\right) \quad (13)$$

Among them, $\alpha_i$ is the attention weight between the concept vector in the $i$th concept set and its short text, $W_1 \in \mathbb{R}^{d_a \times (2u+d)}$ is the weight matrix, $v_1 \in \mathbb{R}^{d_a}$ is the weight vector, $d_a$ is the hyperparameter, and $b_1$ is the bias.

Add the self-attention mechanism in the concept set and calculate the attention to obtain the importance weight of each concept $c_i$ in the whole concept set:

$$\beta_i = \text{softmax}\left(v_2^T \tanh(W_2 c_i) + b_2\right) \quad (14)$$

Among them, $\beta_i$ is the attention weight of the concept vector in the ith concept set, $W_2 \in \mathbb{R}^{d_b \times d}$ is the weight matrix, $v_2 \in \mathbb{R}^{d_b}$ is the weight vector, $d_b$ is the hyperparameter, and $b_2$ is the bias. The attention mechanism gives important concepts greater weight.

After obtaining the attention weights of $\alpha_i, \beta_i$, combine them with Formula (12) to obtain the final attention weight:

$$a_i = \text{softmax}(\gamma \alpha_i + (1 - \gamma)\beta_i) \quad (15)$$

Where $a_i$ is the attention weight of the final $i$-th concept vector, and $\gamma \in [0,1]$ is the weight parameter for adjusting $\alpha_i$ and $\beta_i$.

After obtaining the attention weight of each concept vector, calculate the weight of each concept vector:

$$r = \sum_{i=1}^{m} \alpha_i c_i \quad (16)$$

2.4 Model Training

In this paper, the back propagation algorithm is used to train the network model, and L2 regularization is introduced to avoid the overfitting problem of the network model. L2 regularization can prevent overfitting by adding L2 norms to the loss function as a penalty term, which makes the model fit more in favor of low-dimensional models. Compared with L1 regularization, which causes sparsity problem, L2 regularization can make the coefficient vectors smoother and avoid sparsity problem. In this paper, we optimize the network model by minimizing the cross-entropy loss function:

$$\text{loss} = -\sum_{i=1}^{D} \sum_{j=1}^{C} y_i^j \ln y_i^j + \lambda \parallel \theta \parallel^2 \quad (17)$$

Where $D$ is the training set size, $C$ is *the number of* categories, $y$ is the predicted category, $y'$ is the actual category, and $\lambda \parallel \theta \parallel^2$ is the regularization term.

2.4 Original Local Self-Attention Mechanism

Local Self-Attention mechanism is improved on the basis of Self-Attention this paper, the local self-attention mechanism is introduced to select the data related to the prediction time period. The algorithm does the feature $\tanh$ operation, calculate the attention score of each element; use the softmax function to calculate the attention score $e_t^k$; based on the probability distribution vector and the model input to get the Attention layer output:

$$\begin{aligned} e_t^k &= v_t^e \tan h(w_t^e u_t^k + b_t^e) \\ \beta_t^k &= \text{soft max}(w_t^\beta e_t^k + G_t^k) \\ S_t^k &= \sum_{t=1}^i \beta_t^k (w_t^t u_t^k) \end{aligned} \quad (18)$$

where: $\beta_t^k$ and $G_t^k$ are the probability distribution and Gaussian bias of the attention score at the corresponding moment of the modal component, reflecting the closeness of the connection between that moment and the center position. Gaussian bias $G_t^k$ can again be computed from the following expression.

$$\begin{aligned} G_t^k &= -\frac{(t-Q_t^k)^2}{2\sigma_k^2} \\ \begin{bmatrix} Q_t^k \\ D^k \end{bmatrix} &= I \cdot \text{sigmoid}\left(\begin{bmatrix} q_t^k \\ z^k \end{bmatrix}\right) \\ q_t^k &= v_q^T \tan h(w_q u_t^k) \\ z^k &= v_d^T \tan h(w_d \bar{K}) \end{aligned} \quad (19)$$

In the formula. $Q_t^k$ is the first $k$ A modal component $t$ the predicted center position at the moment of time; $\sigma_k$ can usually be set to $\frac{D^k}{2}$, and $D^k$ is called the window size; the scalar factor $I$ is a real number between 0 and the losing sequence; the $v_q^T, v_d^T, w_q$ and $w_d$ is the weighting coefficient; the $z^k$ is the window selection scalar factor for the modality; the $\bar{K}$ in the original text represents the key values between the semantics.

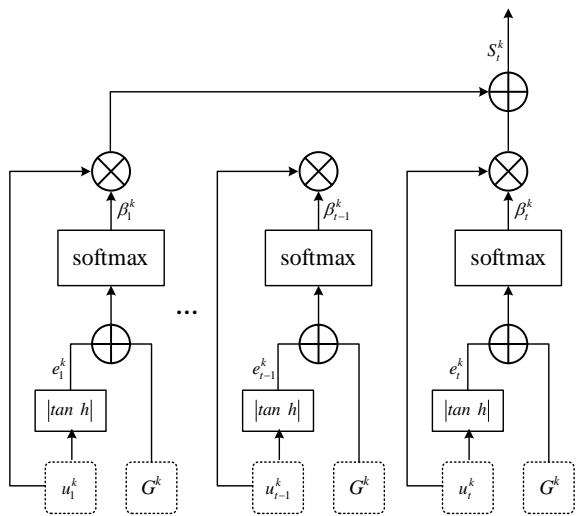

Fig.4 Changing the attention score output

2.5 Improvement of local self-attention mechanism

A large number of related experimental studies have shown that the effect of a single model is often inferior to that of a combined model, which is due to the fact that a single prediction model has a great deficiency in the selection and extraction of effective information. In this paper, we introduce the local self-attention mechanism into the prediction model to better mine the effective features, and select the scalar factor from the attention score formula and window selection, respectively. $z^k 2$ Improvements in individual aspects.

2.5.1 Improvement of the attention score formula

The introduction of the local self-attention mechanism can help the GRU model quickly and accurately select and enhance the vectors with more significant features, but the original local self-attention mechanism cannot fully utilize the useful information. The main reason for the above problem is that after normalization, only the extreme values can be captured and enhanced by local self-attention mechanism. The main reason for the above problem is that after normalization, only the extreme values can be captured and enhanced by the local selfattention mechanism, and the probability score of the very small values, as a feature as important as the very large values, is mostly negative, and the probability of the attention score is very small after the softmax function, which loses a large part of the useful information, as shown in

In this paper, starting from the data characteristics, the formula for calculating the attention score in the local self-attention mechanism is improved, so that both the extremely large and extremely small values can obtain higher attention scores, thus obtaining the same level of attention. The improved attention score formula is

$$e_t^k = |v_t^e \tanh(w_t^e u_t^k + b_t^e)| \quad (20)$$

where: $uu_t^k$ is the value of $t$ The ILSAGRU model input at the moment of time, which contains different feature data at that moment, can be calculated as follows

$$\begin{aligned} u_t^k &= \text{Sigmoid}(W_t U_t^k + b_t^e) \\ U_t^k &= [p_t^k, v_t^k, d_t^k, te_t^k]^T \end{aligned} \quad (21)$$

where: $pp_t^k, v_t^k, d_t^k$ and $te_t^k$ denote the first $k$ sub-modal components at the moment of time $t$; v $v_t^e, w_t^e$ and $W_t$ is the weighting factor; the $b_t^e$ is the bias factor.

2.5.2 Modify the window to select the scalar factor $z^k$

After the variational modal decomposition, different sub-modalities have different center frequencies, it is more reasonable to use different lengths of features to predict the power value in the current time period for the components with different center frequencies, the lower frequency components will capture the trend, and for the higher frequency components, the overly old information will lose its meaning. The model assigns a higher probability to the components close to the predicted time step by Gaussian bias.

The window size $D^k$ is calculated by the window selection scalar factor $z^k$, here the window selects the scalar factor $z^k$ The formula is changed to

$$z^k = U_d^T \tan h\left(\frac{w_d}{\omega_k}\right) \quad (22)$$

where: $U_d^T$ and $w_d$ is the weighting coefficient; the $\omega_k$ is the first $k$ The center frequencies of the sub-modal components, $\omega_k$ The larger the $z^k$ is smaller and vice versa. This not only alleviates the lag problem caused by the GRU network forgetting too quickly and paying too much attention to the power value of the previous moment in the predicted time step, but also pays attention to the characteristics of different center-frequency power series components with different time step lengths and importance, which eliminates the interference of the long history of the information for the current moment prediction.

2.6 Experimental datasets

The method in this paper is tested in two public data sets of different fields to solve the task of sentiment analysis of short texts.

1. Semeval 2017 dataset is the Twitter data set which contains three emotional classifications: positive, neutral and negative.

2. Agnews: Literature [18] A corpus of news articles was obtained through the Internet, including 496,835 classified news articles from more than 2,000 news sources. Only the largest four categories in the corpus were selected from the title and description fields to construct a data set.

3. OHSUMED: The bibliographic classification data set of medical literature published in Literature [19], in which documents with multiple labels are deleted and titles are only used for short text classification.

4. The film review data set and IMDB film review data set published by the Association for Computerized Linguistics (ACL),

each of which contains positive and negative emotional tendencies.

5. TagMyNews: The English news text data set published in reference [20] uses its news headlines as the data set, including seven categories such as politics and art. Corpus statistics are shown in Table 1.

Table 1 Corpora statistics

| Dataset | Number of documents | Number of categories | Average sentence length |
|---|---|---|---|
| Twitter | 49 570 | 3 | 12.3 |
| AGNews | 6 000 | 4 | 18.4 |
| Ohsumed | 7 400 | 23 | 6.8 |
| Movie Review | 9 584 | 2 | 8.6 |
| TagMyNews | 32 549 | 7 | 5.1 |

Table 2 Experimental results of accuracy of each data set

| Model | Twitter | AGNews | Ohsumed | Movie Review | TagMyNews |
|---|---|---|---|---|---|
| CNN | 0.614 | 0.614 | **0.672** | 0.756 | 0.571 |
| ATCNN | 0.648 | 0.683 | 0.342 | 0.773 | 0.594 |
| CNNsLSTMs | 0.653 | 0.679 | 0.336 | 0.768 | 0.574 |
| ATLSTM | 0.659 | 0.684 | 0.347 | 0.771 | 0.584 |
| KCNN | 0.661 | 0.713 | 0.389 | 0.779 | 0.624 |
| KGIAM (ours) | **0.692** | **0.751** | 0.587 | **0.790** | **0.685** |

2.5 Experimental setting

The comparison methods [4, 21-23] for each dataset are as follows: 1) CNN. 2) AT-CNN (Attention-based CNN). 3) CNNs-LSTMs. 4) AT-LSTM (Attention-based LSTM). 5) KCNN (Knowledge Convolutional Neural Network). The pre-training word vector dimension used in the model is 300, the character vector dimension is 50, the concept vector dimension is 100, and the normal distribution random number with standard deviation of 0.1 is randomly initialized by weight. At the same time, all word vectors, character vectors and concept vectors are fine-tuned during training. In the word embedding layer, the pool layer is set with a dropout value of 0.3. Adam optimization method is used to speed up the model training, and the model is trained by 50 samples in each batch.

2.6 Experimental results and analysis

The experimental results of this model and the five comparative models on five public datasets are shown in Table 2. As can be seen from Table 2, the models proposed in this paper achieve better results on datasets in different domains, especially in the Ohsumed and TagMyNews datasets. This is due to the fact that these datasets are constructed by using only the titles of news and articles, the length of the text is too short, and there is a lack of contextual information, which makes the models less effective in the absence of prior knowledge, and the models can generally only be used in the pre-trained word vectors, because the unique nouns or new words may not exist in the pre-trained word vectors when they appear. When there are unique nouns or new words, they may not exist in the pre-trained word vectors, so the model can only use the random initialization method to deal with this problem, which will lead to the poor classification discrimination effect of the model. Both KCNN and the model in this paper achieve excellent classification results because they are integrated with the knowledge graph, which can obtain the a priori knowledge in the text. However, the improvement of this model on the Twitter and Movie Review datasets is not very obvious, mainly due to two reasons: on the one hand, these two datasets are users' blog posts and comments, which may contain certain contextual information; on the other hand, since these two datasets are in the direction of sentiment categorization, the model obtains the features of the sentiment words that have a greater impact on the classification effect, and the recognition of the sentiment words may not require much a priori knowledge, so the model has a better classification effect. On the other hand, since the two datasets are for sentiment classification, the model obtains sentiment word features that have a greater impact on the classification effect, and recognizing sentiment words often may not require a lot of a priori knowledge, and the pre-trained word vectors can express the sentiment features well. Therefore, the improvement between KCNN and this model on these two datasets is not obvious.

We discuss three examples of short text conceptualization, in which the first two are news classification, the first short text is classified as business news, the second is classified as science and technology news, and the third is classified as medical literature. This short text is classified as parasite-related literature. As can be seen from the first two news short texts, words appearing in the short texts may express wrong meanings by using word vectors. For example, Apple is an Apple company in the short texts, and the word vectors are likely to express the meaning of fruits, so it is necessary to attract people to the knowledge base to obtain the prior knowledge of words, and the concept sets obtained from the knowledge base often have concepts unrelated to the short texts. For example, Hip Pop's concept of music style has nothing to do with business news, so it attracts people's attention. It calculates the attention of each concept, short text and concept set, and reduces the weight of those concepts that have nothing to do with short text and concept set to prevent it from affecting the classification effect of the model. For the short text of the third medical classification, it can be seen that there may be special words in a certain field in the short text, such as the word Ascaris lumbricoides in the text, which are often not found in the pre-training word vector, so the classification ability of the model will become worse. Because of the combination of knowledge base, the model in this paper can obtain the prior knowledge of the short text, thus solving this problem.

From the above, it can be seen that the model in this paper has a certain degree of improvement over the KCNN model in each domain dataset. As the model in this paper borrows from the Transformer structure, it uses two-way GRU combined with multi-Head self-attention to encode short text, and does Attention computation inside the input sequence to give higher Attention weight to the key words, so as to obtain the connection between the sequences, and at the same time, the multi-Head part puts each head computed by self-attention into the KCNN model, which is the same as the KCNN model. At the same time, the multi-Head part uses different linear transformations for each head computed by self-attention to learn different word relationships. The model in this paper also uses the attention mechanism to encode the input concept set, by calculating the attention weights of the short text and the concepts in the concept set, and assigning higher weights to the concepts that are closely related to the short text, it can solve the noisy concepts in the input of the KCNN model to a certain extent, and make important concepts get higher weights, so the text model has a certain degree of superiority.

Table 3 Impact of parameter γ on the model

| Parameter | Twitter | AGNews | Ohsumed | Movie Review | TagMyNews |
|---|---|---|---|---|---|
| $\gamma=0.00$ | 0.681 | 0.742 | 0.550 | 0.782 | 0.675 |
| $\gamma=0.25$ | 0.692 | 0.750 | 0.561 | 0.790 | 0.682 |
| $\gamma=0.50$ | 0.680 | 0.751 | 0.587 | 0.787 | 0.685 |
| $\gamma=0.75$ | 0.682 | 0.745 | 0.572 | 0.781 | 0.677 |
| $\gamma=1.00$ | 0.683 | 0.742 | 0.561 | 0.779 | 0.672 |

The results are shown in Table 3. From Table 3, we can see that the model works best when $\gamma = 0.25$ or $\gamma = 0.50$, and the selection of specific parameters depends on the dataset. When the parameter $\gamma$ is set to 0 or 1, the model is less effective. This is due to the fact that when $\gamma = 1$, the model ignores the importance of each concept in relation to the set of concepts, which leads to a decrease in the performance of the model. When $\gamma = 0$, the model ignores the semantic similarity of concepts with respect to short texts, in which case concepts that are not related to short texts may be given a larger weight, which affects the classification effect of the model.

3 Conclusion

In this paper, we propose a model that combines knowledge graphs and self-attention for short text categorization. The model first selects each word using the information gain method, and then encodes text context using bi-directional GRU and incorporates improved self-attention to focus on word relations. It leverages a priori knowledge for short texts, addressing contextual information limitations. In addition, we improve the calculation formula for attention scores in the local self-attention mechanism to ensure that words with different frequencies of occurrence in the text can receive higher attention scores, thus achieving an equal level of importance. Experimental results on five different domain datasets validate the method's feasibility and effectiveness.